\documentclass[conference]{IEEEtran}

\usepackage{amsmath,amssymb,amsfonts}
\usepackage{algorithmic}
\usepackage{graphicx}
\usepackage{textcomp}
\usepackage{xcolor}
\usepackage{hyperref}
\usepackage{todonotes}
\usepackage{booktabs}
\usepackage{tabularx}
\usepackage{cleveref}
\usepackage{todonotes}
\usepackage{float}
\usepackage{dblfloatfix}
\usepackage{subcaption}
\usepackage{caption}

\usepackage[style=ieee,natbib=true]{biblatex}
\renewcommand{\bibfont}{\footnotesize}
\DeclareMathOperator*{\minimize}{minimize}

\newtheorem{definition}{Definition}

\begin{document}

\title{
Collaborative Filtering under Model Uncertainty
}
\author{\IEEEauthorblockN{\textbf{Robin M. Schmidt}\IEEEauthorrefmark{1}\thanks{\IEEEauthorrefmark{1}Equal contribution. Author ordering determined randomly by coinflip.}}
\IEEEauthorblockA{\textit{Department of Computer Science} \\
\textit{University of T{\"u}bingen}\\
T{\"u}bingen, Germany \\
\texttt{rob.schmidt@student.uni-tuebingen.de}}
\and
\IEEEauthorblockN{\textbf{Moritz Hahn}\IEEEauthorrefmark{1}}
\IEEEauthorblockA{\textit{Department of Computer Science} \\
\textit{University of T{\"u}bingen}\\
T{\"u}bingen, Germany \\
\texttt{moritz.hahn@student.uni-tuebingen.de}}
}

\maketitle

\begin{abstract}
In their work, \citet{DeanRR20} create a model to research \emph{recourse} and \emph{availability} of items in a recommender system. We used the definition of \emph{predictive multiplicity} by \citet{MarxCU19} to examine different variations of this model, using different values for two model parameters. Pairwise comparison of their models show, that most of these models produce very similar results in terms of \emph{discrepancy} and \emph{ambiguity} for the \emph{availability} and only in some cases the availability sets differ significantly.

All analysis code for our experiments and models can be found on GitHub: \\ \url{https://github.com/SirRob1997/collaborative_filtering}.
\end{abstract}

\begin{IEEEkeywords}
reachability, collaborative filtering, recommender system, multiplicity, ambiguity, discrepancy
\end{IEEEkeywords}

\section{Introduction}
In toady's society, recommendation systems have a huge influence on how individuals explore and experience information \citep{DeanRR20}. Generally, they are applied in a broad variety of domains including media (e.g.~videos or music), product recommendations or travel and real estate. This raises some interesting questions like: ``How easy can a user be pigeonholed by their viewing history?'' or ``How does a recommender system encode bias that limits the availability of content?''. These are some of the main question which inspired the work by \citet{DeanRR20} which form the basis of our contributions by providing the respective models and definition frameworks. 

\section{Model by \citet{DeanRR20}}

\subsection{Problem Setting}

\begin{figure*}
\begin{subfigure}{.5\textwidth}
  \centering
  \includegraphics[width=\textwidth]{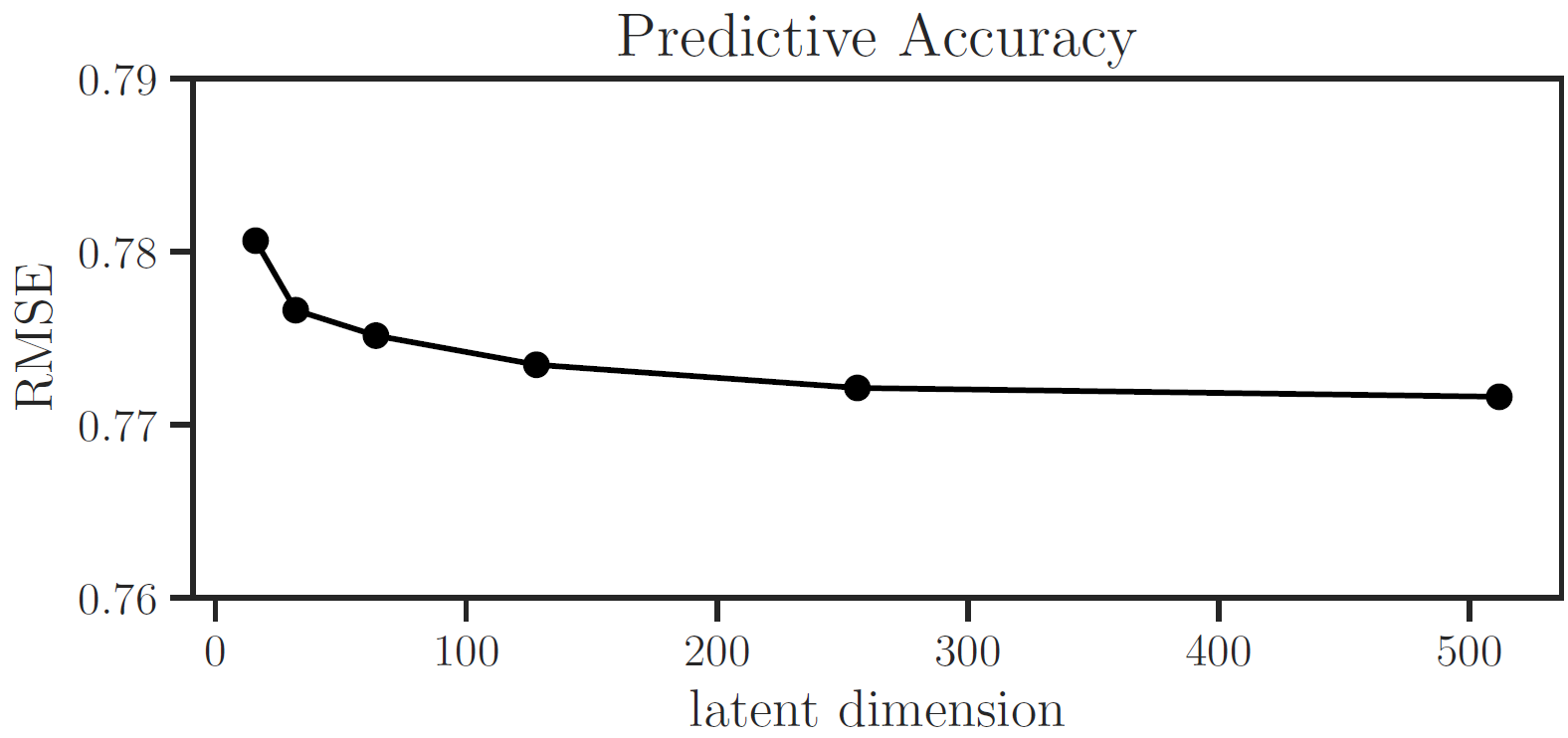}
    \caption{MovieLens dataset}
    \label{fig:ml_pred}
\end{subfigure}
\begin{subfigure}{.5\textwidth}
  \centering
  \includegraphics[width=\textwidth]{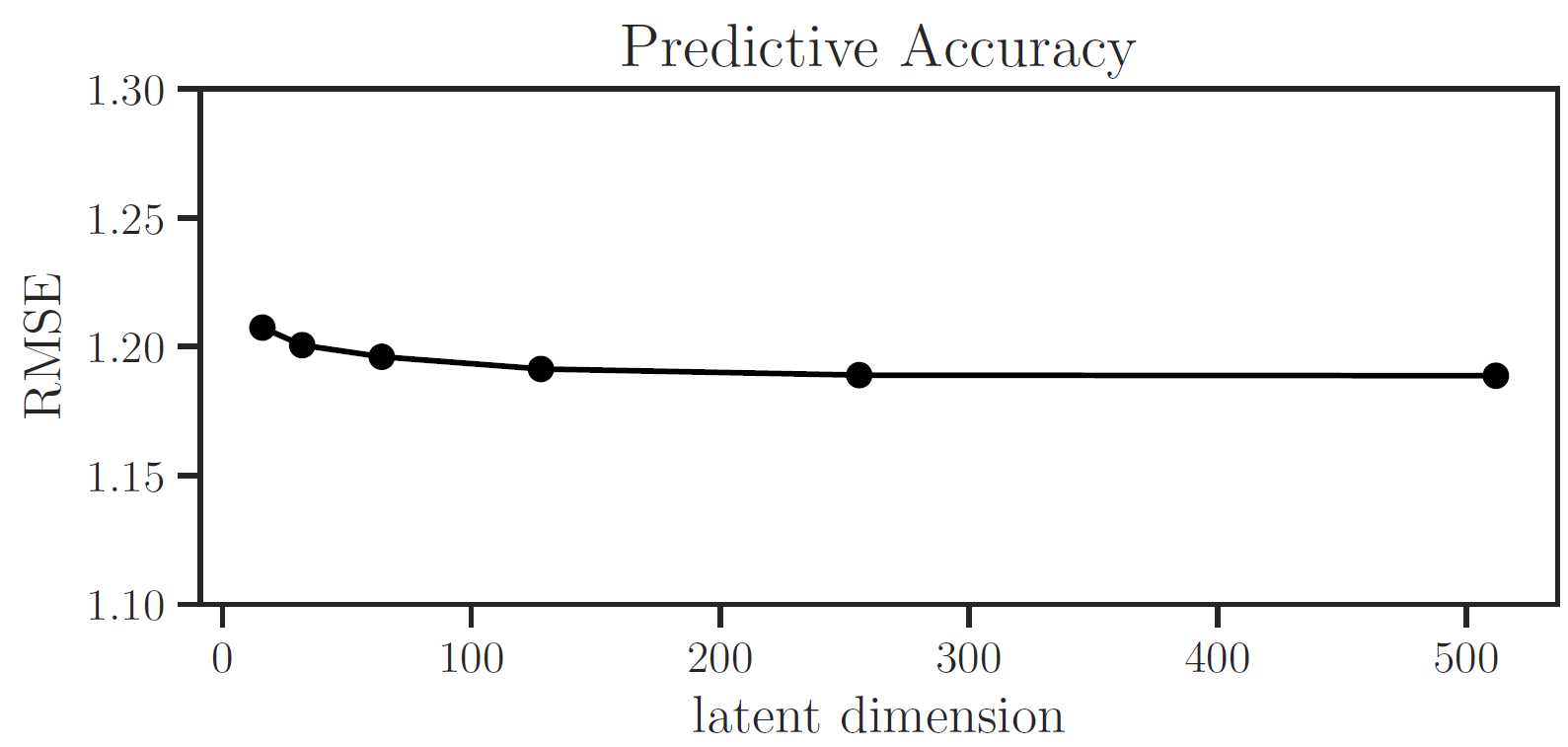}
    \caption{LastFM dataset}
    \label{fig:fm_pred}
\end{subfigure}
\caption{Test RMSE ($y$-axis) of the matrix factorization models with varying latent space dimensions ($x$-axis) on the MovieLens dataset (left) and the LastFM dataset (right): \citep{DeanRR20}}
\label{fig:pred}
\end{figure*}

For their model \citet{DeanRR20} define a recommender system as a collection of $n$ users $u$ and $m$ items $i$. A rating for a user-item combination is then denoted as $r_{u i} \in \mathcal{R} \subseteq \mathbb{R}$. The observed ratings $\Omega_u$ for a user  are stored inside a rating history sparse vector $\mathbf{r}_{u} \in \mathcal{R}^{m}$ with defined values at $\Omega_u$ and $0$ at every other position.
This allows the recommender system to make decisions following a policy based on this sparse vector denoted as $\pi\left(\mathbf{r}_{u}\right)$ which yields a subset of items. 

Based on these constraints, \citet{DeanRR20} define an item $i$ to be reachable from user $u$ if there is a modification to the rating history  $\mathbf{r}_{u}$ that results in item $i$ being recommended to user $u$. With that, they define the whole reachability problem as
\begin{equation}
\label{eq:1}
\begin{aligned}
& \displaystyle{\minimize_{\mathbf{r} \in \mathcal{M}\left(\mathbf{r}_{u}\right)} \quad \operatorname{cost} \left(\mathbf{r} ; \mathbf{r}_{u}\right) } \\
& \operatorname{subject}\, \operatorname{to}\; \; i \in \pi(\mathbf{r})
\end{aligned}
\end{equation}
utilizing a modification set $\mathcal{M}\left(\mathbf{r}_{u}\right) \subseteq \mathcal{R}$ which describes the possibilities of modifications to their respective rating history and the difficulty of making these changes in $\operatorname{cost} \left(\mathbf{r} ; \mathbf{r}_{u}\right)$. Intuitively, the \texttt{cost} function can correlate to the number of needed changes or how far these changes are away from the current preferences of the user \citep{DeanRR20}. 

Moreover, \citet{DeanRR20} focus on linear preference models which use a user vector $\mathbf{p}_u$ and an item vector $\mathbf{q}_i$ in combination with item bias $b_i$, user bias $c_u$ and over all bias $\mu$ to yield a predicted user rating $\widehat{r}_{u i}$ as
\begin{equation}
\widehat{r}_{u i}=\mathbf{q}_{i}^{\top} \mathbf{p}_{u}+b_{i}+c_{u}+\mu\ .
\end{equation}

When incorporating \textit{matrix factorization} with user and item representations as factors lying in a latent space of size $d$ we can represent the factors as $P \in \mathbb{R}^{n \times d}$ and $Q \in \mathbb{R}^{m \times d}$. Together with the regularizer $\Gamma$, this yields
\begin{equation}
    \displaystyle{\minimize_{P,Q} \sum_{u} \sum_{i \in \Omega_{u}}\left(r_{u i}-\mathbf{p}_{u}^{\top} \mathbf{q}_{i}\right)^{2}+\Gamma(P, Q)}
\end{equation}
for fitting the model where \citet{DeanRR20} use the $\ell_2$ regularization on user and item factors.

When defining the \texttt{cost} function, instead of modeling it as the penalty on \textit{change from existing ratings}, \citet{DeanRR20} penalize the \textit{change from predicted ratings}. For edits on observed items (history edits) it is defined as 
\begin{equation}
\operatorname{cost}_{\mathrm{hist}}\left(\mathbf{r} ; \mathbf{r}_{u}\right)=\left\|\mathbf{r}-\mathbf{r}_{u}\right\|
\end{equation}
while edits on the recommended items (reactions) are defined as
\begin{equation}
\operatorname{cost}_{\text {react }}\left(\mathbf{r} ; \mathbf{r}_{u}\right)=\left\|\mathbf{r}_{\pi\left(\mathbf{r}_{u}\right)}-\widehat{\mathbf{r}}_{\pi\left(\mathbf{r}_{u}\right)}\right\|\ .
\end{equation}
\subsection{Recourse and Availability}

Further, \citet{DeanRR20} define \textit{recourse} and \textit{availability} as they are respectively important for understanding how a user's preferences limited the reachable content and how available a certain item is inside the recommender system. Hence, they are defined as:

\begin{definition}[recourse]
The amount of recourse available to a user $u$ is defined as the percentage of unseen items that are reachable, i.e. for which \Cref{eq:1} is feasible. The difficulty of recourse is defined by the average value of the recourse problem over all reachable items $i$.
\end{definition}

\begin{definition}[availability]
The availability of items in a recommender system is defined as the percentage of items that are reachable by some user. 
\end{definition}

\subsection{User Cold-Start}
When a new user enters the recommender system, he has no prior rating history from which to predict preferences from. This process is known as the User Cold-Start problem and common strategies for recommender systems include presenting items which are most likely to be rated highly or be most informative about the respective users' preferences \citep{DeanRR20}. With the definition of recourse, \citet{DeanRR20} define the onboarding set not only for its contribution to the model accuracy but also its provided amount of recourse. 

\subsection{Sufficient Conditions for Top-$N$}

\begin{figure*}[!htbp]
\centering
\begin{subfigure}{.3\textwidth}
  \centering
  \includegraphics[width=\textwidth]{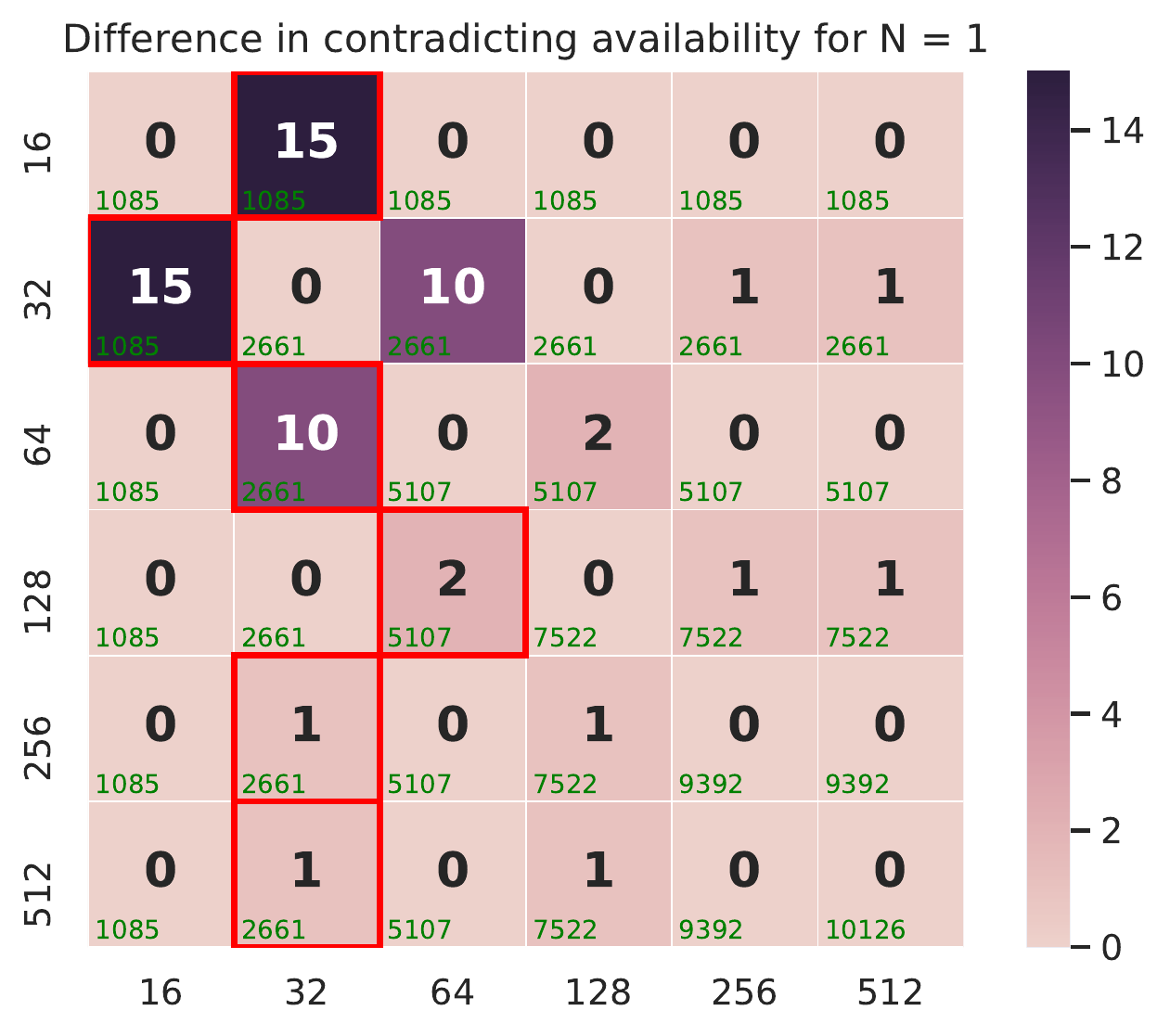}
    \caption{Top-$1$ Recommender System}
    \label{fig:ml_av_1}
\end{subfigure}
\hfill
\begin{subfigure}{.3\textwidth}
  \centering
  \includegraphics[width=\textwidth]{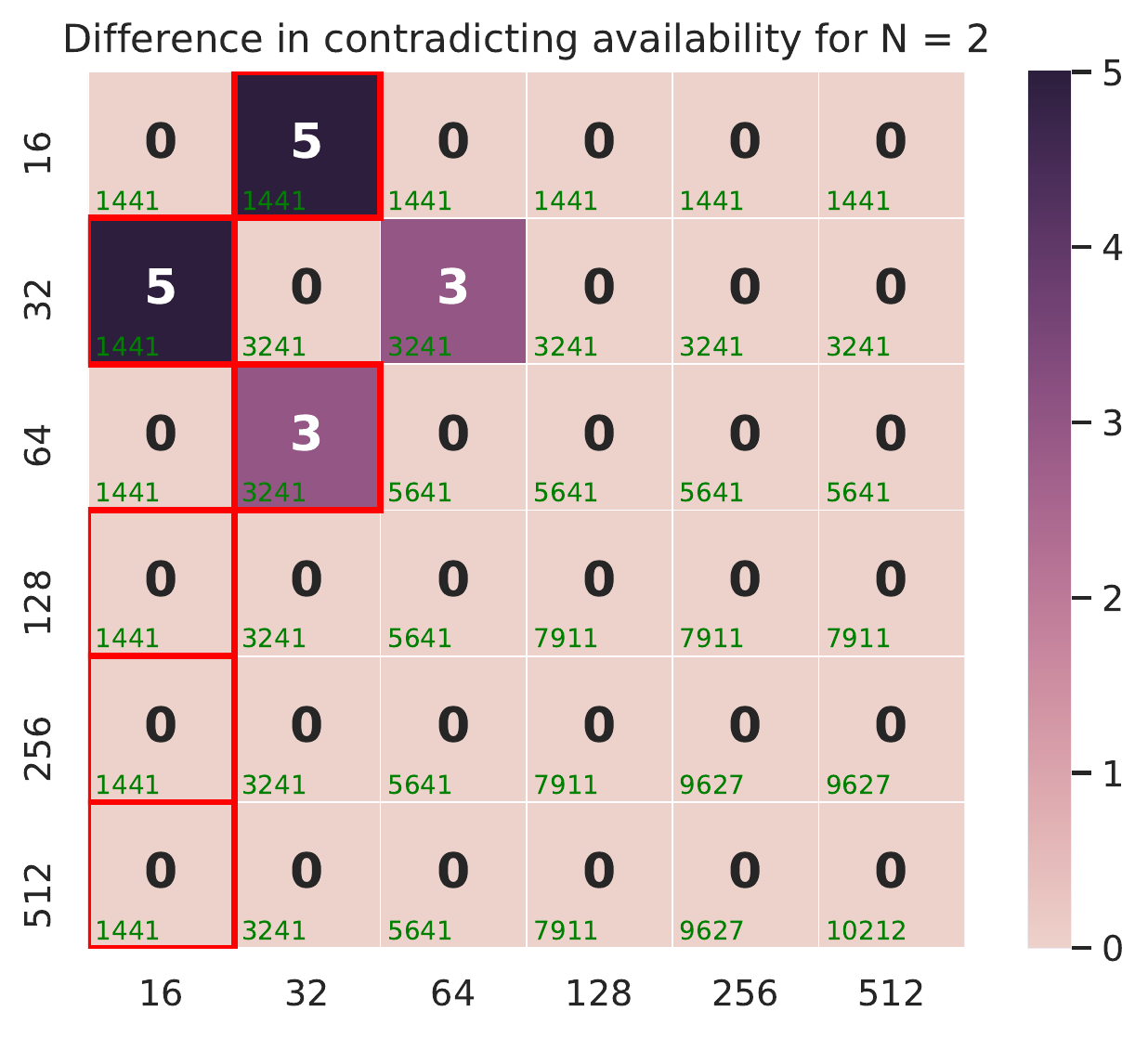}
    \caption{Top-$2$ Recommender System}
    \label{fig:ml_av_2}
\end{subfigure}
\hfill
\begin{subfigure}{.3\textwidth}
  \centering
  \includegraphics[width=\textwidth]{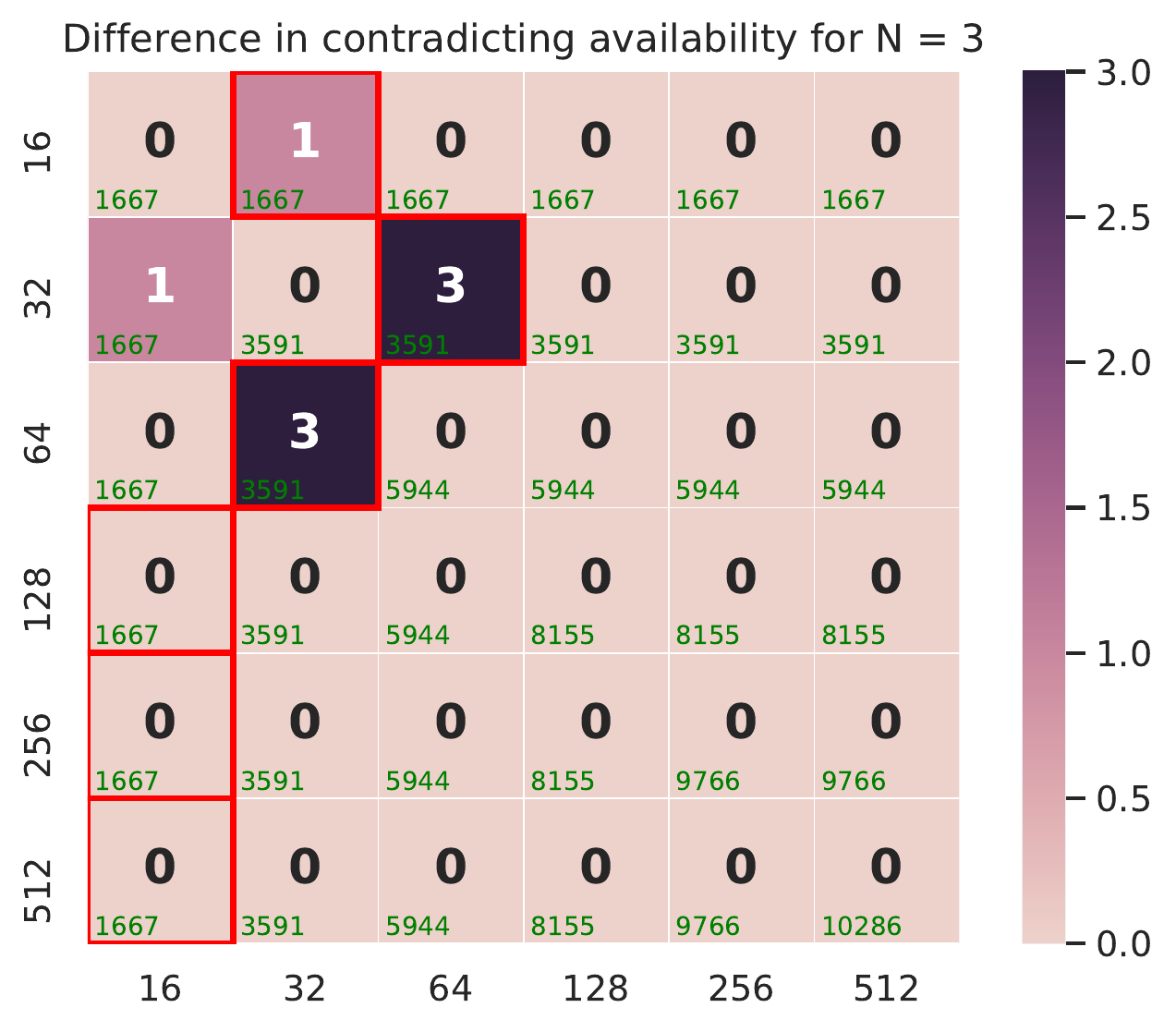}
    \caption{Top-$3$ Recommender System}
    \label{fig:ml_av_3}
\end{subfigure}
\\[5pt]
\begin{subfigure}{.3\textwidth}
  \centering
  \includegraphics[width=\textwidth]{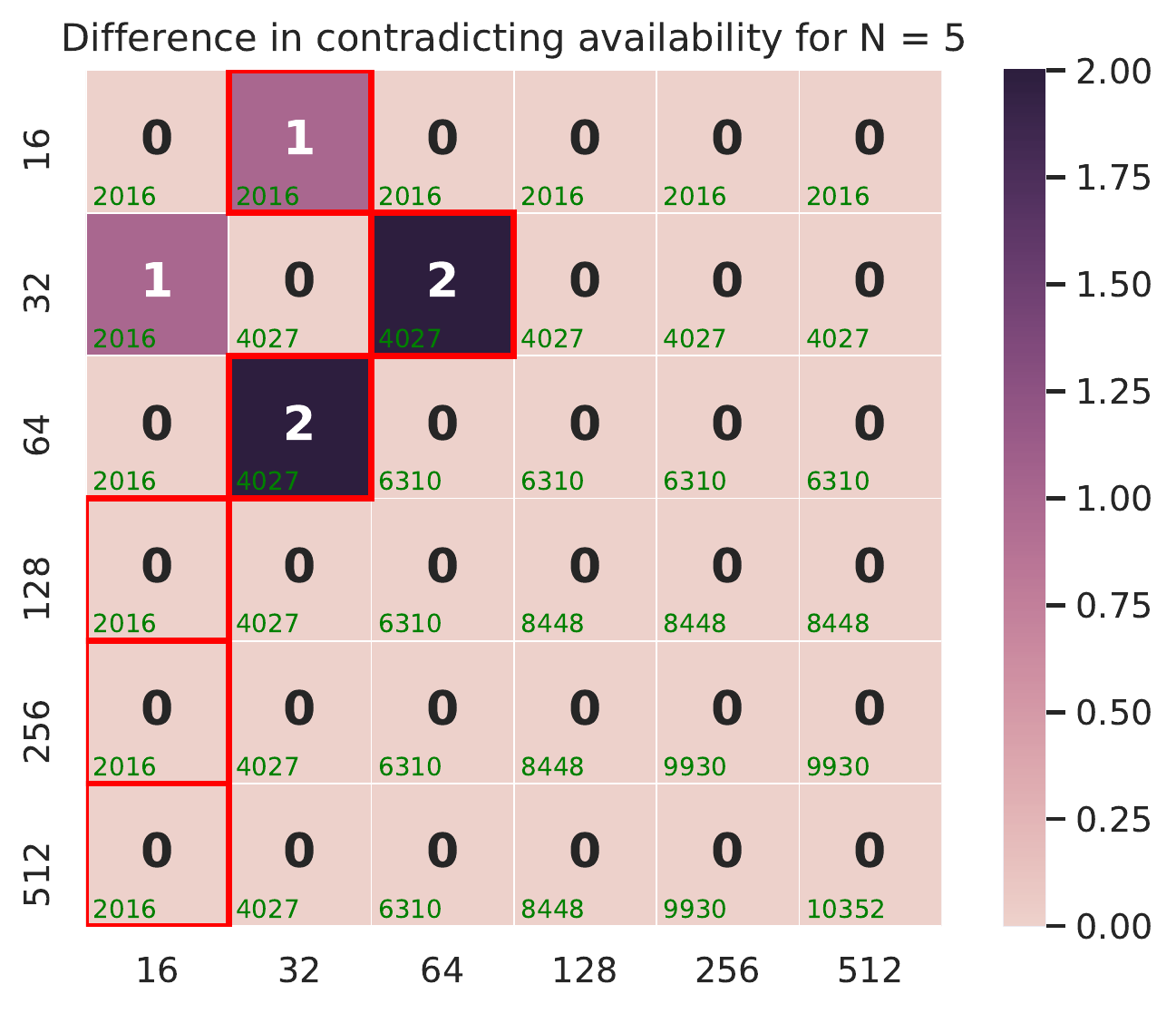}
    \caption{Top-$5$ Recommender System}
    \label{fig:ml_av_5}
\end{subfigure}
\hfill
\begin{subfigure}{.3\textwidth}
  \centering
  \includegraphics[width=\textwidth]{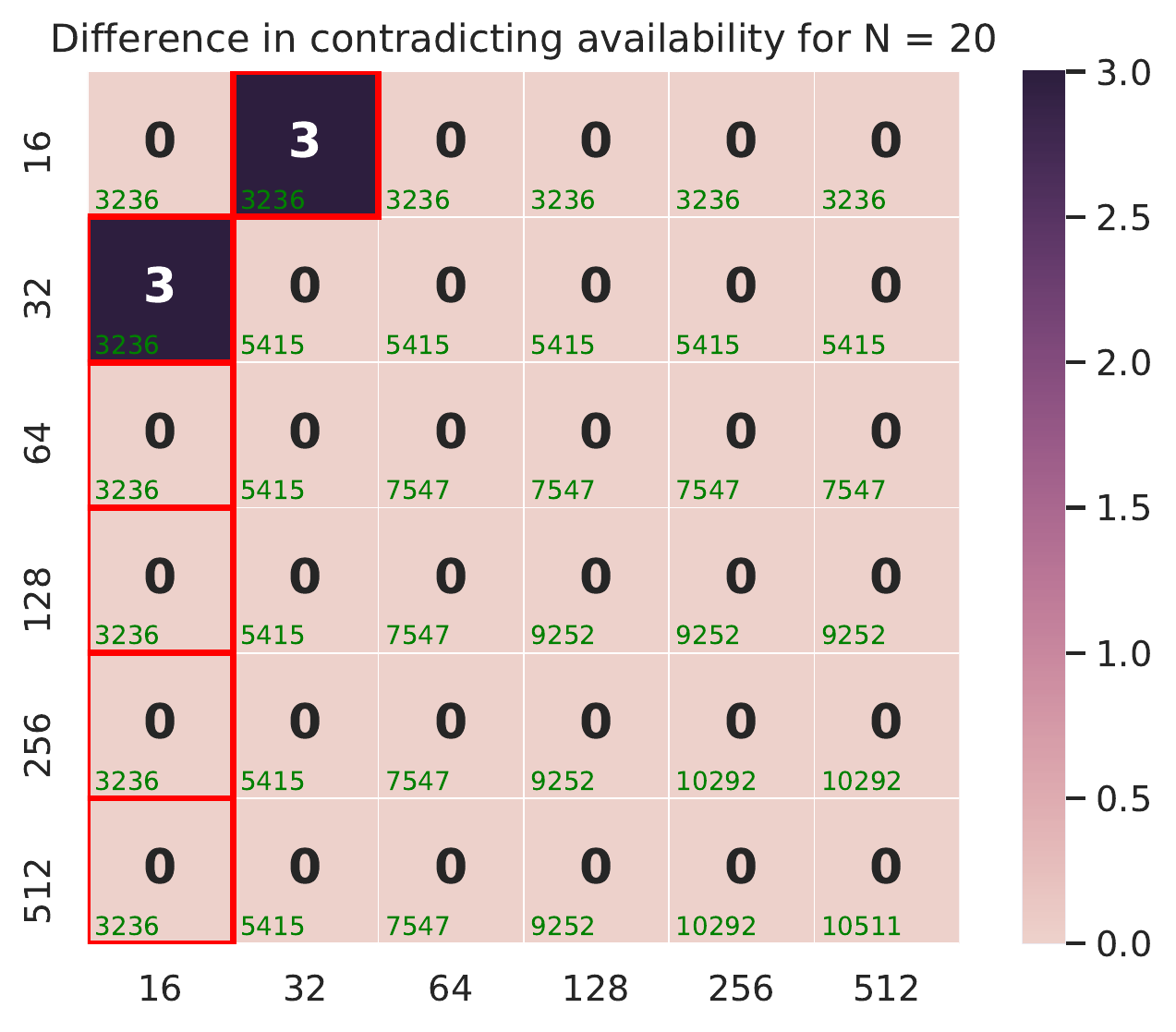}
    \caption{Top-$20$ Recommender System}
    \label{fig:ml_av_20}
\end{subfigure}
\hfill
\begin{subfigure}{.3\textwidth}
  \centering
  \includegraphics[width=\textwidth]{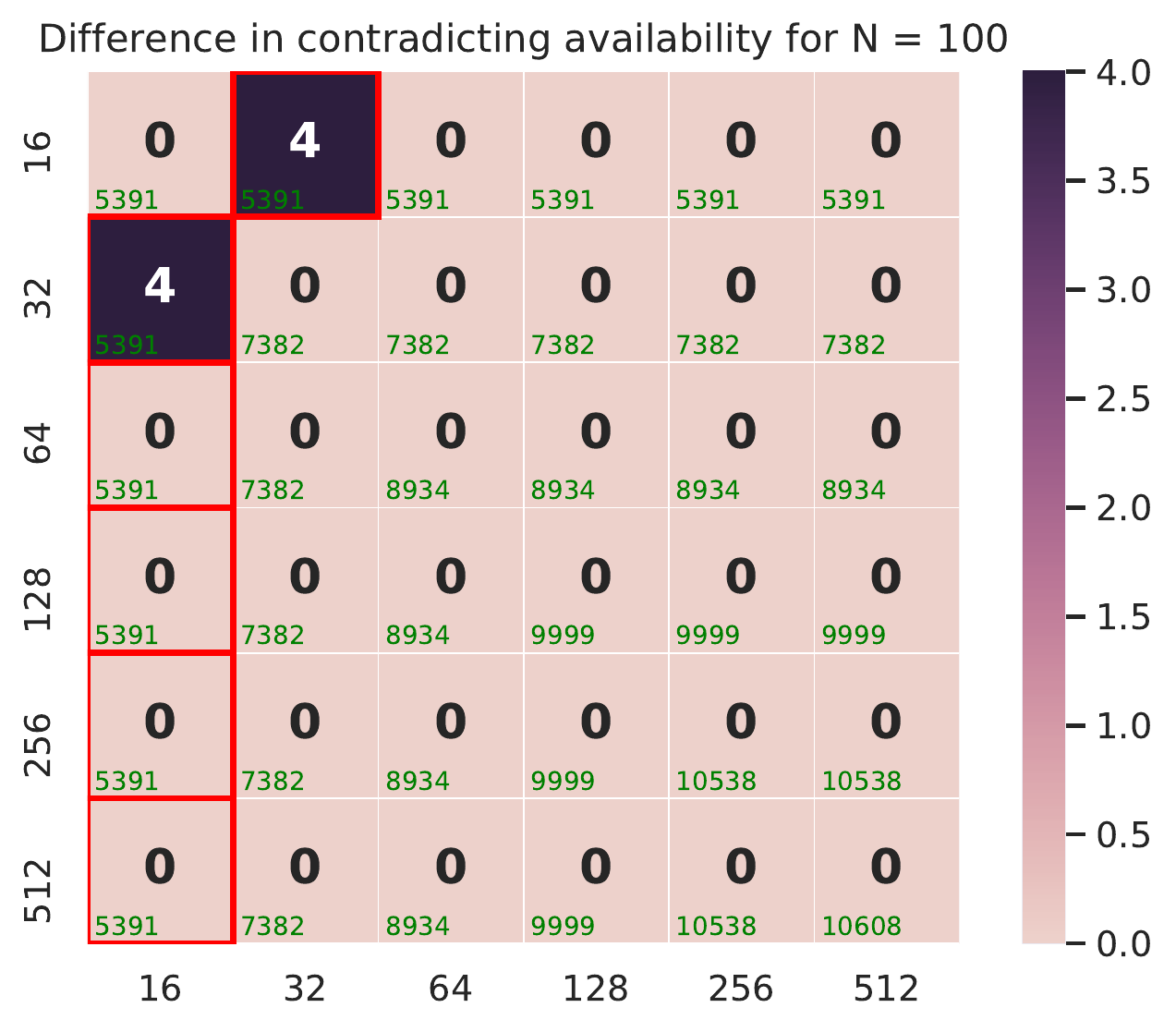}
    \caption{Top-$100$ Recommender System}
    \label{fig:ml_av_100}
\end{subfigure}
\caption{Discrepancy of availability on the MovieLens dataset comparing the available items $\mathbb{Y}$ of any baseline model ($y$-axis) with the available items $\mathbb{X}$ of any model in the $\epsilon$-level set  ($x$-axis) for varying latent space size $d$. The content of each cell is the amount of elements in the difference set $\mathbb{Z} = \mathbb{Y} \setminus \mathbb{X}$ where $|\mathbb{Y}| \leq |\mathbb{X}|$ and $ \mathbb{Z} = \mathbb{X} \setminus \mathbb{Y}$ otherwise. Final discrepancy (row-wise maximum) for the baseline model and the size of available items of the smaller set are highlighted in red and green.} 
\label{fig:ml_av}
\end{figure*}

The recommender system described by \citet{DeanRR20}, a Top-$1$ recommender system, only ever recommends one item at a time. Since most real world applications involve serving several items at once, to model reality more closely, \citet{DeanRR20} expand this system to recommend $N$ items at the same time instead, creating a Top-$N$ recommender system with $N > 1$. 

\citet{DeanRR20} define an item-region for the Top-$N$ case, when $i \in \pi(\textbf{p}; \Omega)$ as follows:

\begin{equation}
\mathcal{P}_i = \textbf{p} : (\textbf{q}_i - \textbf{q}_j )^{\top} \textbf{p} > 0 
\end{equation}
with all but at most $N$ items $j \in
/ \Omega$ .

Although, according to \citet{DeanRR20}, this region is contained in the latent space, generally of relatively small dimensions, the description depends on the number of items which in general will be quite large. While linear for $N = 1$, for $N > 1$, the description for each region requires $\mathcal{O} (m^N)$ linear inequalities, becoming expensive very quick, even for small values of $N$.

\subsubsection{Sufficient Condition for Availability}

To bypass those computational concerns, \citet{DeanRR20} show that the full description of the Region $\mathcal{P}_i$ is not necessary, but instead finding any point in the latent space $\mathbf{v} \in \mathbb{R}^d$ that satisfies $\mathbf{v} \in \mathcal{P}_1$ is sufficient. They propose a sampling approach to determine the availability of an item with a complexity of only $\mathcal{O}(m^2d \log (m))$.

\citet{DeanRR20} call items $i$ that are inside an item-region defined via sampling \textit{aligned-reachable}, which is a lower bound on the availability of items, yielding an underestimate of the availability of items in a system.

Using the aligned-reachable condition as a generic model audit, \citet{DeanRR20} propose an item-based audit algorithm with $\Omega = \emptyset$ and an increased value for $N$. The model audit counts the number of aligned-unreachable items, returning a lower bound on the overall availability of items. The model audit can also be used to propose constraints or penalties on the model during training.

\subsubsection{Sufficient Condition for Recourse}

As user recourse inherits the same computational problems as described for availability, \citet{DeanRR20} continue with the sampling perspective to test feasibility. This yields a lower bound on the amount of recourse available to a user, based on their specific rating history and the allowable actions. They show, that items who are aligned-reachable are also reachable by users, implying that item availability implies recourse for any user with control over at least $d$ ratings whose corresponding item factors are linearly independent.

\citet{DeanRR20} conclude, that user recourse follows from the ability to modify ratings for a set of diverse items, and immutable ratings ensure the reachability of some items, potentially at the expense of others.

\subsection{Experimental Demonstration}

\begin{figure*}[ht]
\centering
\begin{subfigure}{.3\textwidth}
  \centering
  \includegraphics[width=\textwidth]{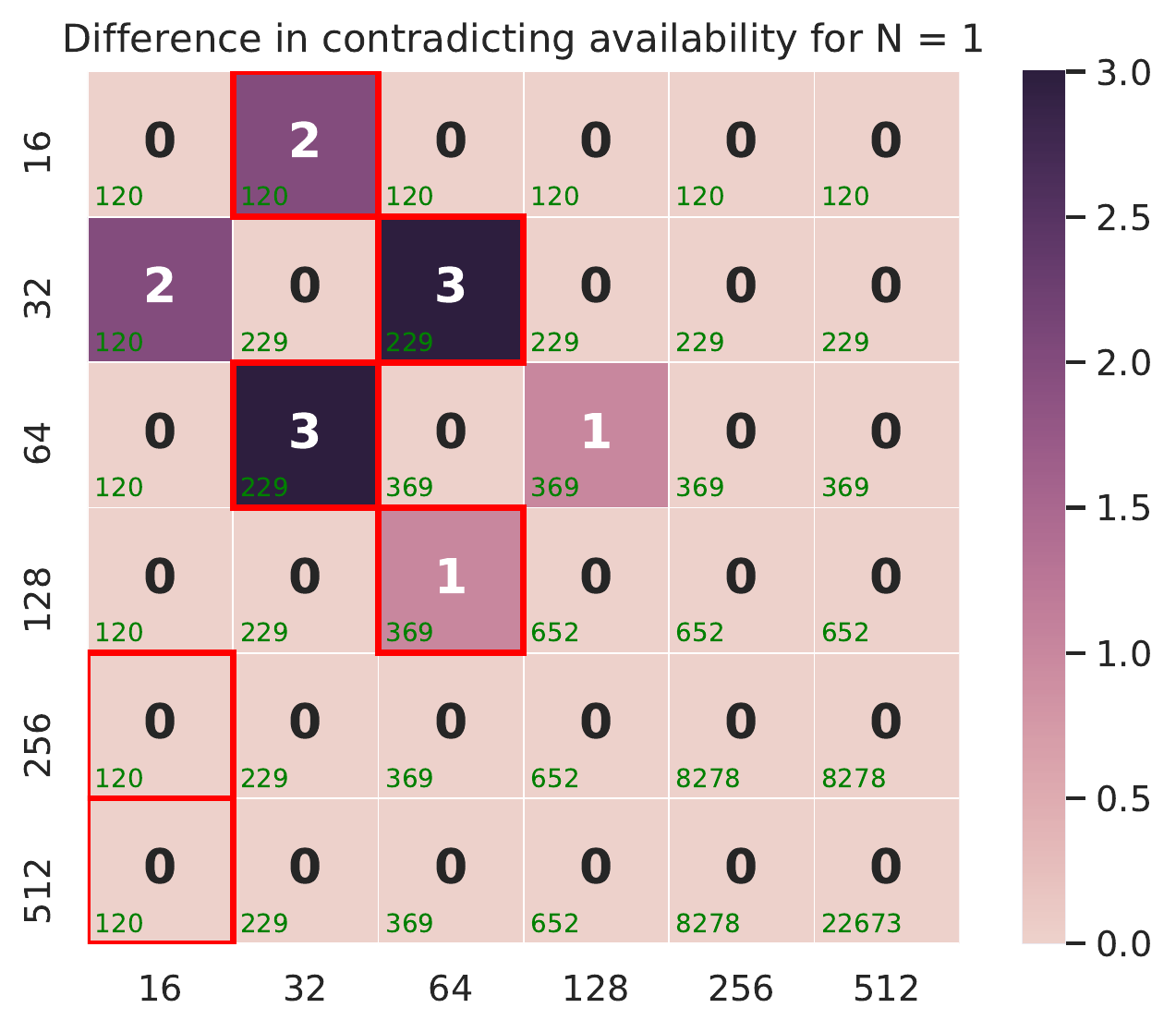}
    \caption{Top-$1$ Recommender System}
    \label{fig:fm_av_1}
\end{subfigure}
\hfill
\begin{subfigure}{.3\textwidth}
  \centering
  \includegraphics[width=\textwidth]{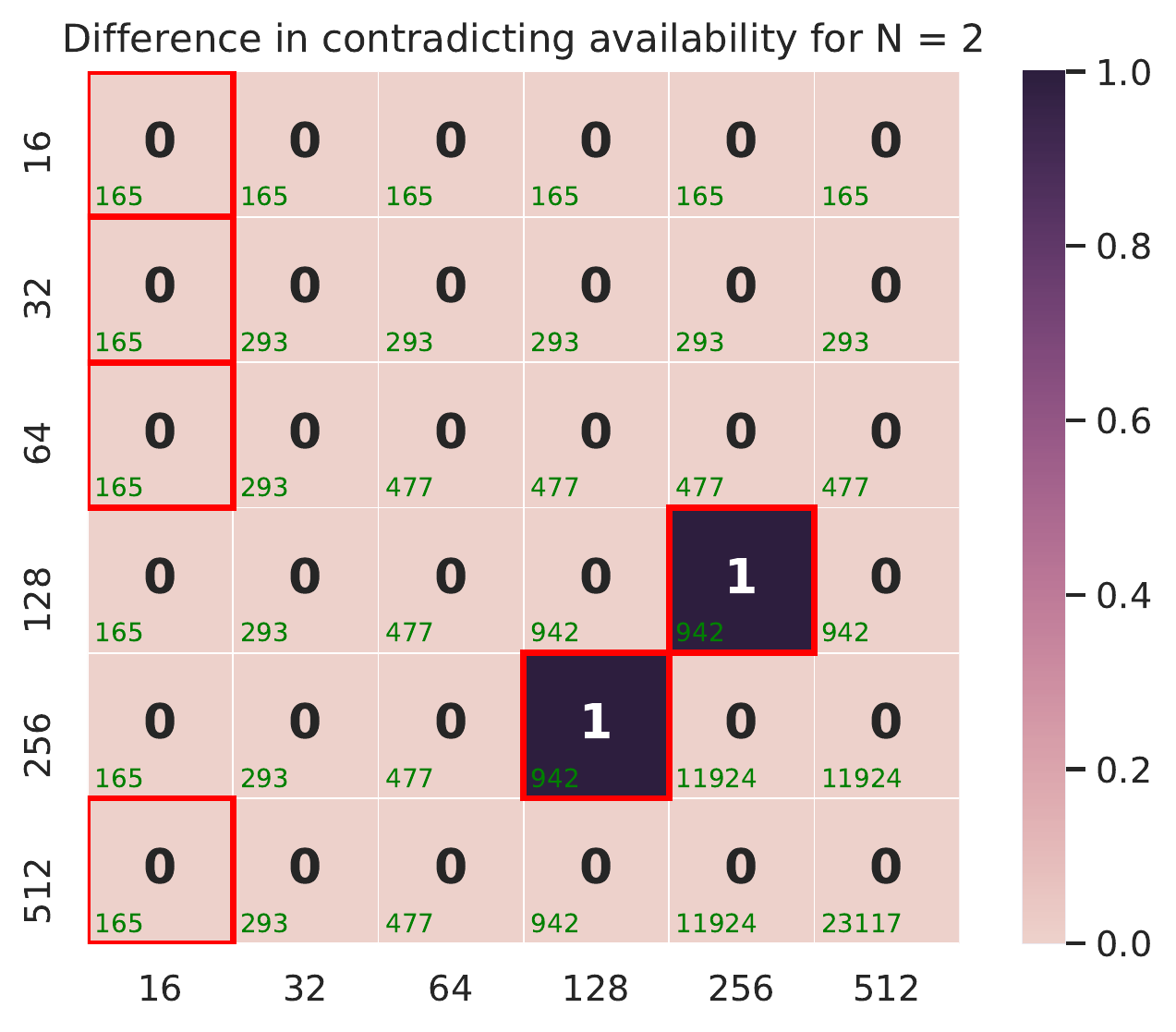}
    \caption{Top-$2$ Recommender System}
    \label{fig:fm_av_2}
\end{subfigure}
\hfill
\begin{subfigure}{.3\textwidth}
  \centering
  \includegraphics[width=\textwidth]{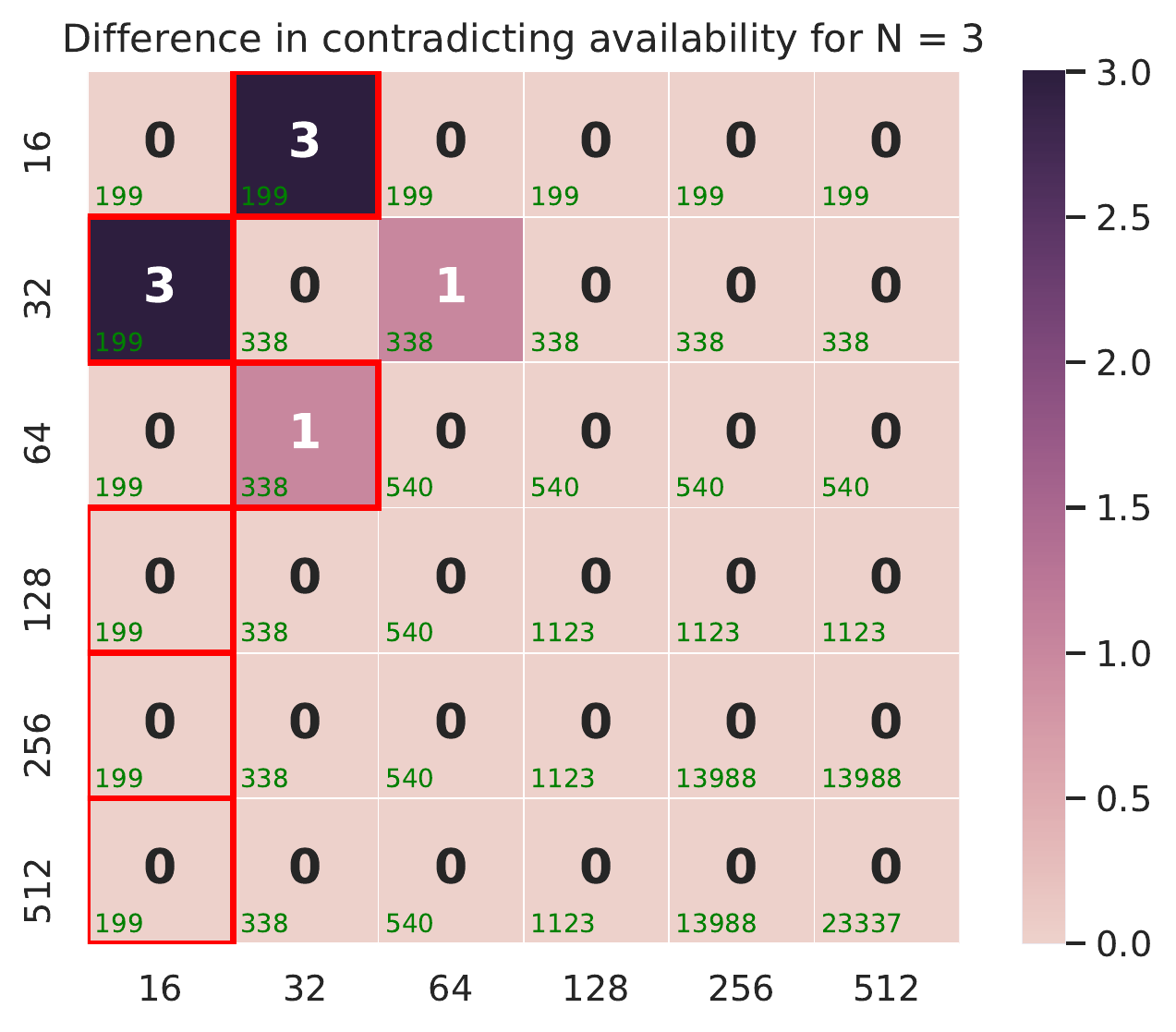}
    \caption{Top-$3$ Recommender System}
    \label{fig:fm_av_3}
\end{subfigure}
\\[5pt]
\begin{subfigure}{.3\textwidth}
  \centering
  \includegraphics[width=\textwidth]{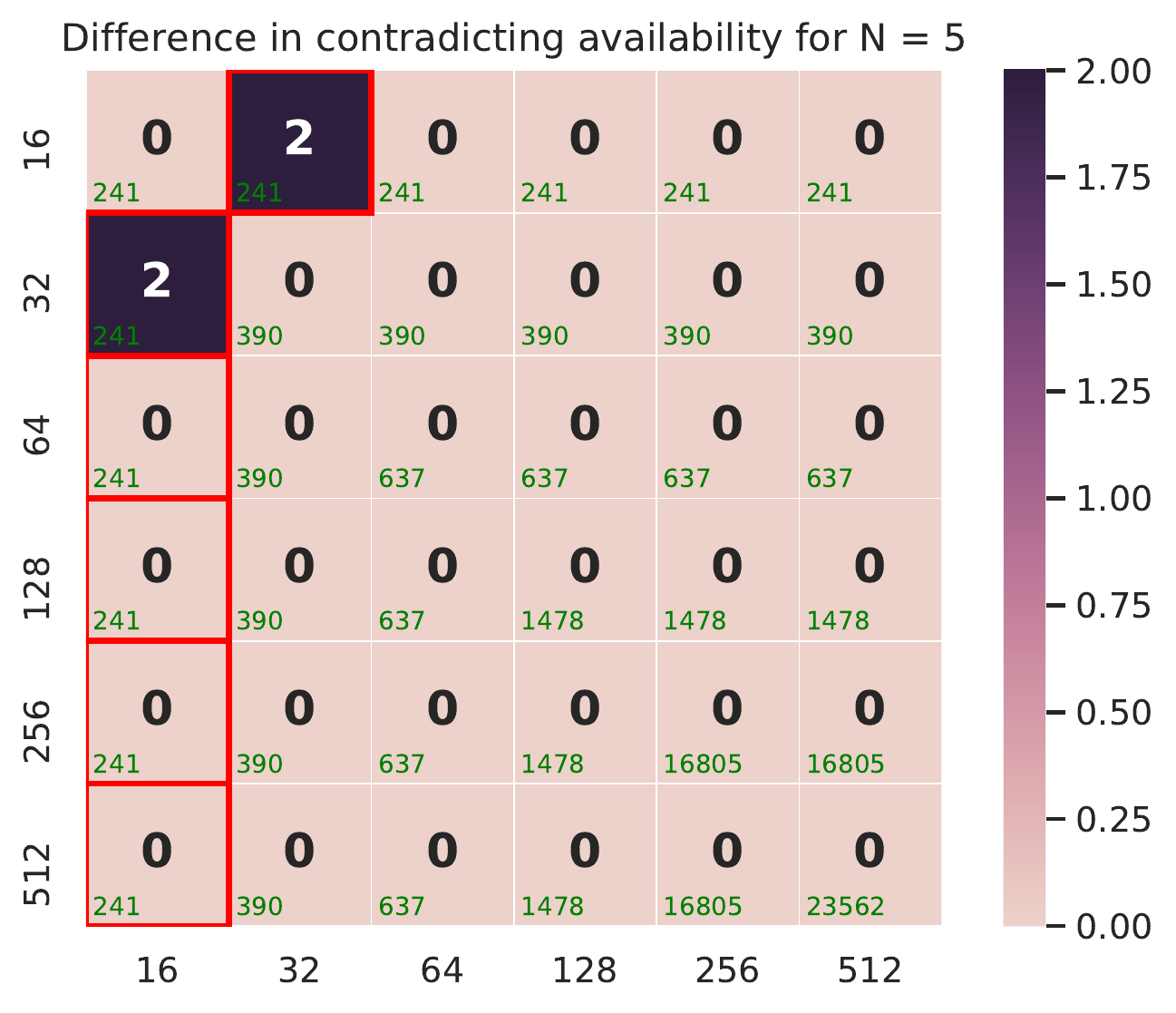}
    \caption{Top-$5$ Recommender System}
    \label{fig:fm_av_5}
\end{subfigure}
\hfill
\begin{subfigure}{.3\textwidth}
  \centering
  \includegraphics[width=\textwidth]{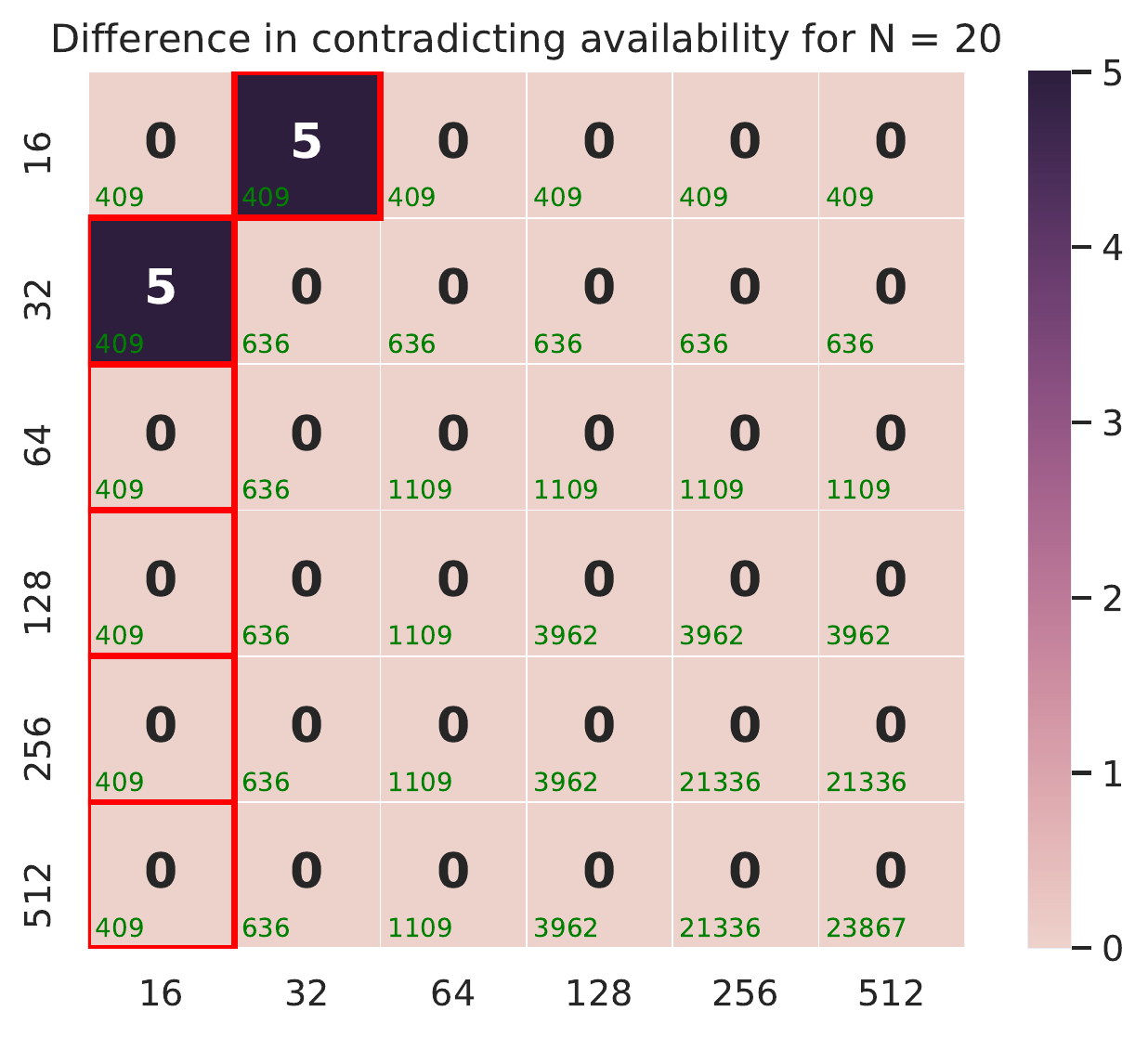}
    \caption{Top-$20$ Recommender System}
    \label{fig:fm_av_20}
\end{subfigure}
\hfill
\begin{subfigure}{.3\textwidth}
  \centering
  \includegraphics[width=\textwidth]{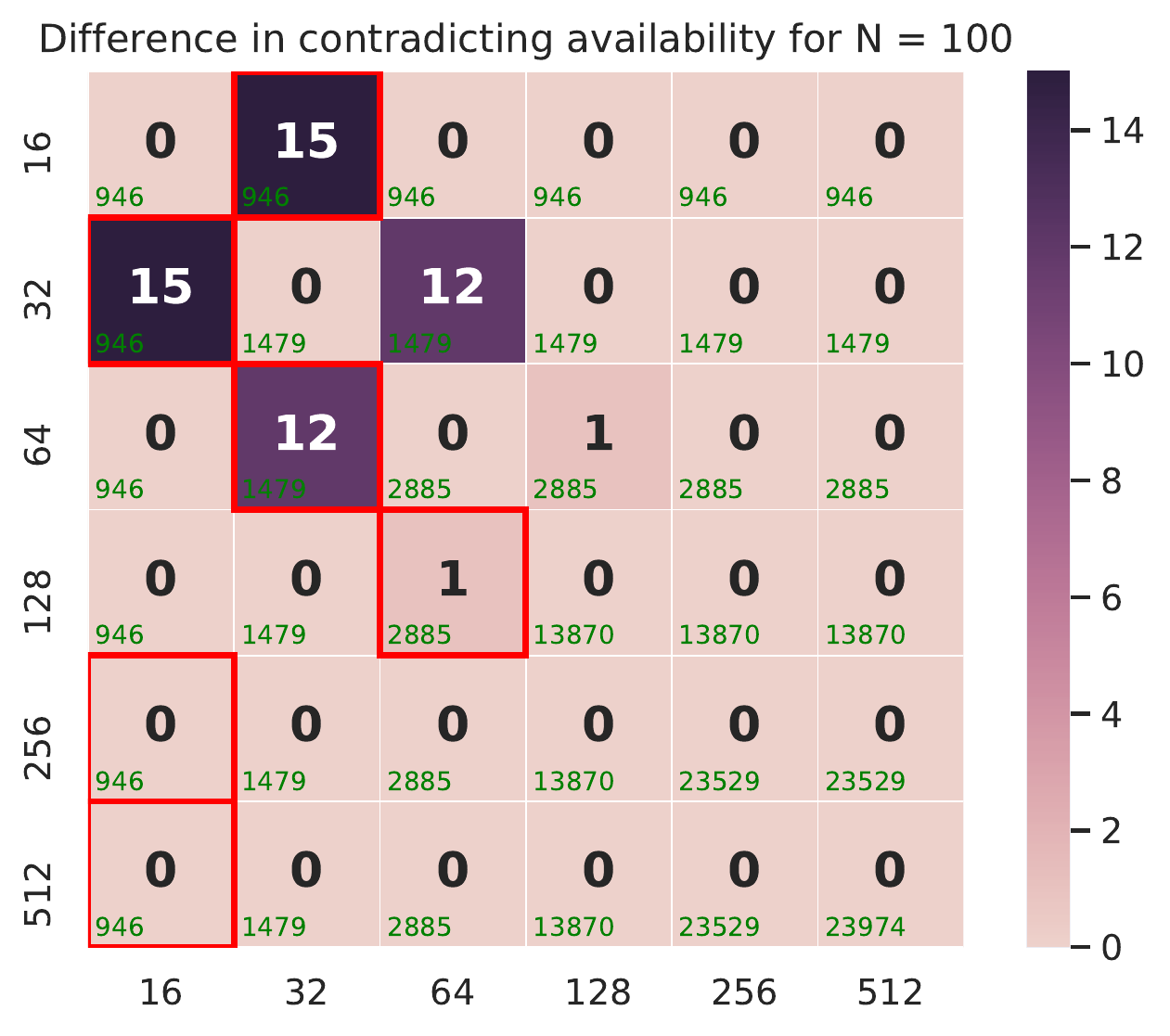}
    \caption{Top-$100$ Recommender System}
    \label{fig:fm_av_100}
\end{subfigure}
\caption{Discrepancy of availability on the LastFM dataset comparing the available items $\mathbb{Y}$ of any baseline model ($y$-axis) with the available items $\mathbb{X}$ of any model in the $\epsilon$-level set  ($x$-axis) for varying latent space size $d$. The content of each cell is the amount of elements in the difference set $\mathbb{Z} = \mathbb{Y} \setminus \mathbb{X}$ where $|\mathbb{Y}| \leq |\mathbb{X}|$ and $\mathbb{Z} = \mathbb{X} \setminus \mathbb{Y}$ otherwise. Final discrepancy (row-wise maximum) for the baseline model and the size of available items of the smaller set are highlighted in red and green.}

\label{fig:fm_av}
\end{figure*}

\citet{DeanRR20} demonstrate how the proposed analyses can be used for auditing and interpreting characteristics of a matrix factorization model. They use the MovieLens 10M dataset \citep{Harper2016}, a common benchmark for evaluating rating predictions and the method described by \citet{RendleZK19} in their recent work on baselines for recommender systems. The methods did match those presented by \citet{RendleZK19} and reproduced their reported accuracies. Models of a variety of latent dimension ranging from $d = 16$ to $d = 512$ were examined. Additionally, they conducted a similar set of experiments on the LastFM dataset \citep{Bertin-Mahieux2011}, which yielded similar results. Those results were included in Appendix B of the original paper.

After performing the item-based audit previously described, \citet{DeanRR20} found, that for larger values of $d$ as well as $N$, the number of items that are aligned-reachable is significantly higher. Baseline reachability is especially low for small values of $d$. At the same time, they found that unavailable items do have systematically lower popularity, while popularity alone does not determine reachability.

\subsubsection{System Recourse for Users}

For this, \citet{DeanRR20} used continuous ratings instead of rounded to the nearest $.5$ step, since it was easier for the model to work with, also only the $1000$ most rated items were chosen to significantly reduce the computation time needed. This does produce a small overestimation because of popularity bias but was still considered a good approximation.

When allowing history edits, for a growing number of items in the history, \citet{DeanRR20} recognised two distinct shapes in the recourse curve. For all values of $d$, there first was a sharp increase in the available recourse, which did level after a while for each value of $d$. This can be explained by two factors: The sharp increase is determined by the limiting effect of the projection, which rises continuously. The leveling of effect on the other hand is determined by the baseline item-reachability. \citet{DeanRR20} note at this point that while higher complexity and therefore a larger number of latent dimensions provides a larger amount of recourse, lower complexity lets the model reach the maximum faster.

When considering fixed ratings and allowing no history edits, \citet{DeanRR20} were able to come up with the following conclusions:

\begin{enumerate}
\item The amount of recourse is actually bigger for a lower number of latent dimensions
\item For a small history length, more recourse is available
\item When rating random items, the available recourse is bigger, compared to rating only the recommended items
\end{enumerate}

This does not contradict the previous results, as using a fixed history does eliminate the availability of additional recourse and only concerns the anchor points. It is worth noting, that the advantages of additional recourse seem to outweigh the disadvantages of the anchor points for large histories and more latent dimensions.

\subsubsection{Recourse difficulty}

At last, \citet{DeanRR20} did examine the cost of recourse over all users for a single item. A Top-$1$ recommender system was used to once again reduce the computational burden of computing the exact set $\mathcal{P}$. The \textit{cost} was posed as the size of the difference between the user input and the predicted ratings. In two trial runs, the cost of recourse was determined for a set of $20$ random items as well as for the set of the $20$ highest rated items. Here, \citet{DeanRR20} was able to arrive at the following findings:

\begin{enumerate}
\item The cost does not increase, but the amount of recourse is lower for a larger number of latent dimensions.
\item The cost was actually lower for random, then for Top-$20$ 
\end{enumerate}

The experimental demostration ended with the conclusion, that future work should more carefully examine methods for constructing recommended sets that trade-off predicted ratings with measures like diversity under the lens of user recourse.

\section{Drawbacks of the Dean et al. Model}

\begin{figure*}[ht]
\centering
\begin{subfigure}{.49\textwidth}
  \centering
  \includegraphics[width=\textwidth]{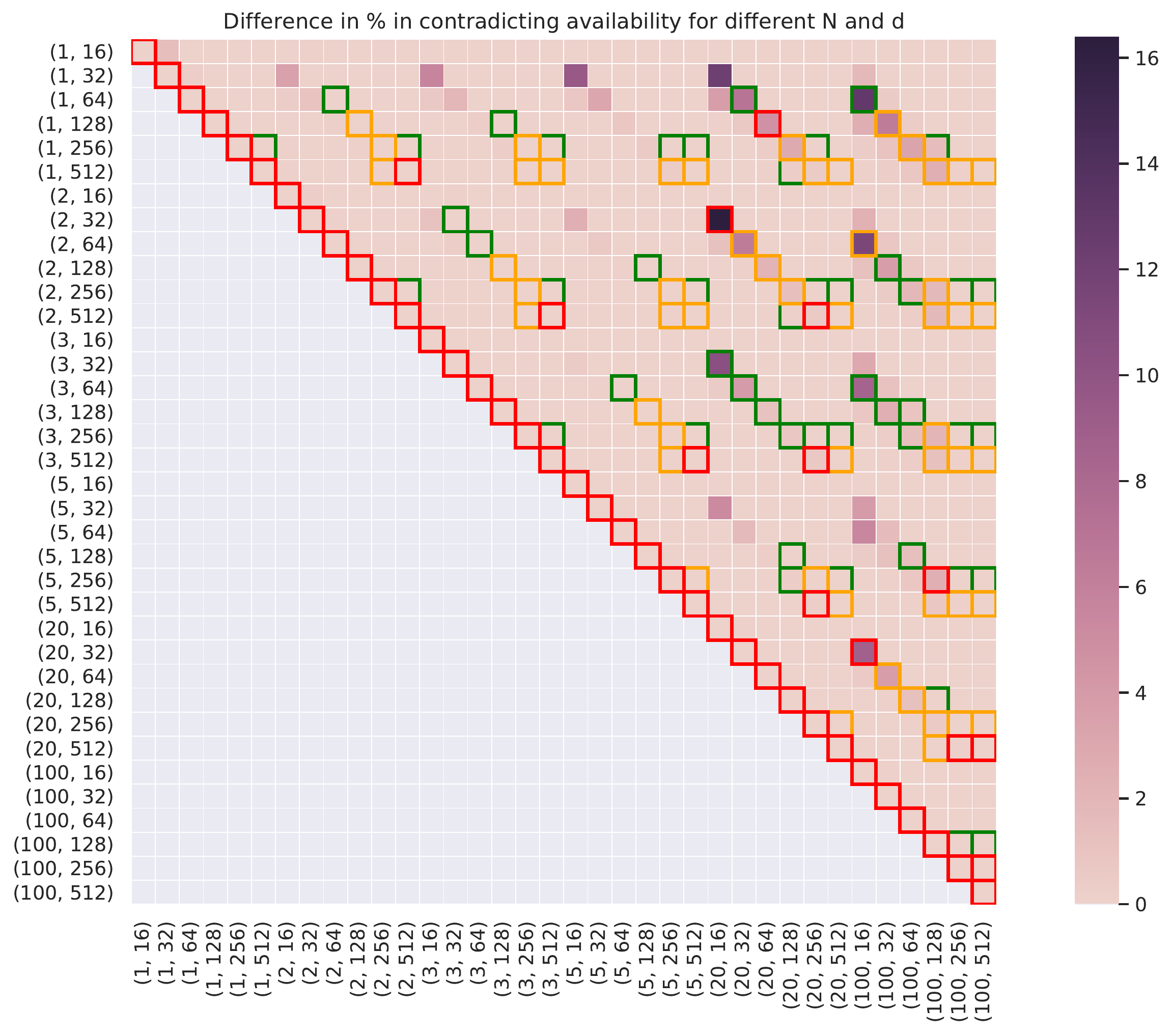}
    \caption{MovieLens dataset}
    \label{fig:ml_big}
\end{subfigure}
\begin{subfigure}{.49\textwidth}
  \centering
  \includegraphics[width=\textwidth]{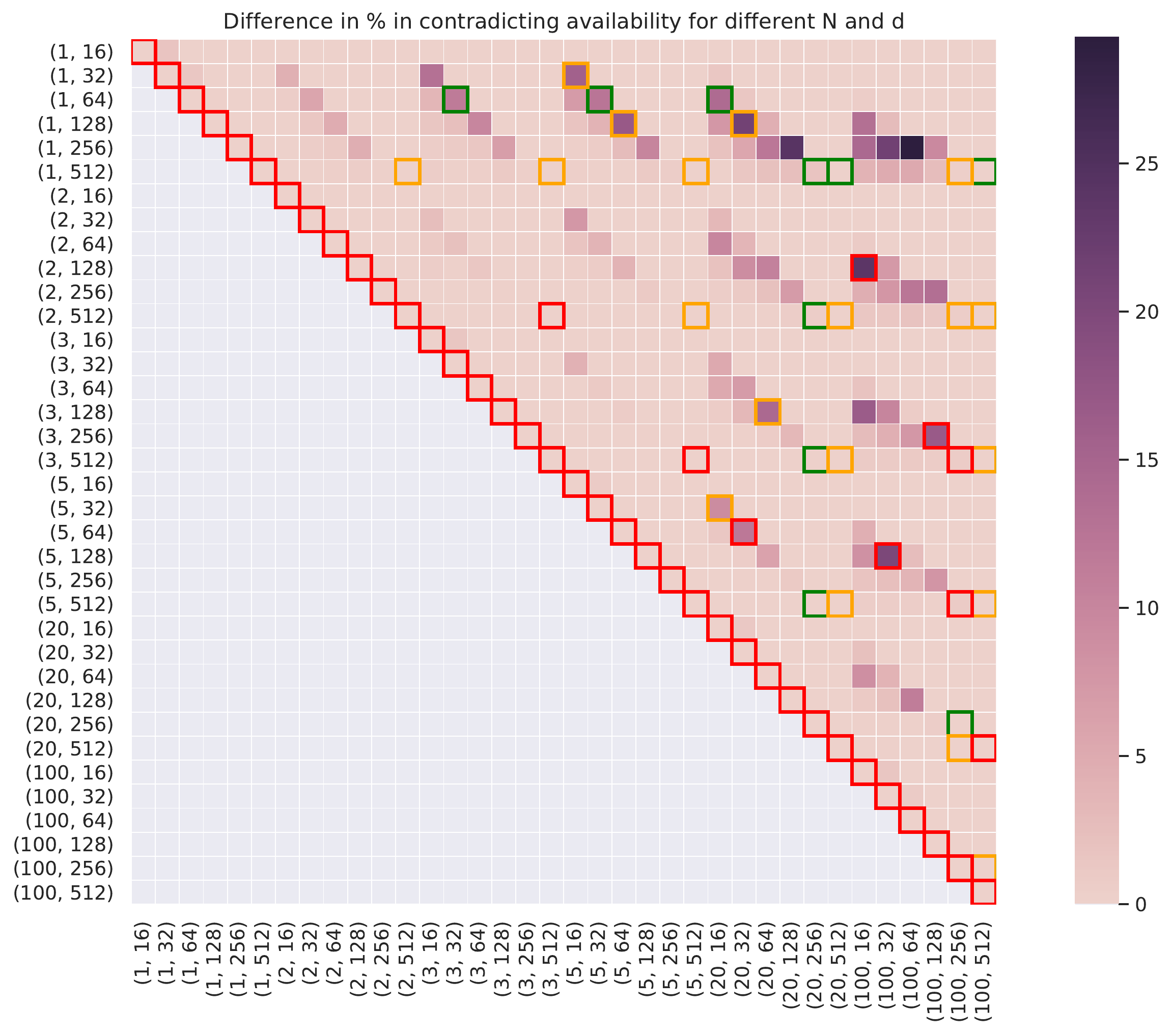}
    \caption{LastFM dataset}
    \label{fig:fm_big}
\end{subfigure}
\caption{Percentage of availability discrepancy on the MovieLens (left) and LastFM dataset (right) comparing the available items $\mathbb{Y}$ of any baseline model ($y$-axis) with the available items $\mathbb{X}$ of any model in the $\epsilon$-level set  ($x$-axis) for varying Top-$N$ recommender systems and latent space size $d$ denoted as $(N,d)$. The content of each cell is the amount of elements in the difference set $\mathbb{Z} = \mathbb{Y} \setminus \mathbb{X}$ where $|\mathbb{Y}| \leq |\mathbb{X}|$ and $\mathbb{Z} = \mathbb{X} \setminus \mathbb{Y}$ otherwise. Comparisons where the set size difference $|Z|$ is less than $\{10\%, 5\%, 1\%\}$ are marked with $\{\text{green}, \text{orange}, \text{red} \}$ borders.}
\label{fig:pred_big}
\end{figure*}

The method proposed by \citet{DeanRR20} has a number of drawbacks, which are described briefly and left for further research. These drawbacks include the user cold start problem, popularity biases, filter bubbles and human-model interactions. 

\subsubsection{User Cold Start}

While proposing onboarding sets to provide additional recourse, rather than focusing on model accuracy, \citet{DeanRR20} do not provide any demonstration of how the onboarding set plays a potential role in availability and recourse.

\subsubsection{Popularity bias}

\citet{DeanRR20} find that, the differences in availability of items does, to some extend, relate to their general popularity or unpopularity in the training data. According to \citet{Steck11}, this seems to be a phenomenon in recommender systems in general which amongst other things reproduces undesirable demographic biases. \citet{DeanRR20} provide no further explanation on how to combat this problem.

\subsubsection{Filter Bubbles}

The model used by \citet{DeanRR20} solely provides a reachability criteria which is based on the possibility of a user reaching a specific item. However, it does not provide any predictions if a specific user in a real-world scenario will actually reach the specific item or not. The possibility of recourse, for example, does not fix the problem of filter bubbles, as it merely provides the means to do so, but the user also has to use these means. Therefore, further research would be warranted to examine if the cost function proposed models actual user behaviour or if it needs to be fundamentally changed to not provide a false appearance of fairness.

\subsubsection{Human-Model Interactions}

Lastly, there are untapped possibilities of future work to examine the interactions between users and models as the models evolve over time and the user behaviour might be influenced by the model at the same time. \citet{DeanRR20} note, that this path likely would lead towards understanding phenomena like filter bubbles.

\section{Model by \citet{MarxCU19}}

\begin{figure*}[ht]
\centering
\begin{subfigure}{.4\textwidth}
  \centering
  \includegraphics[width=\textwidth]{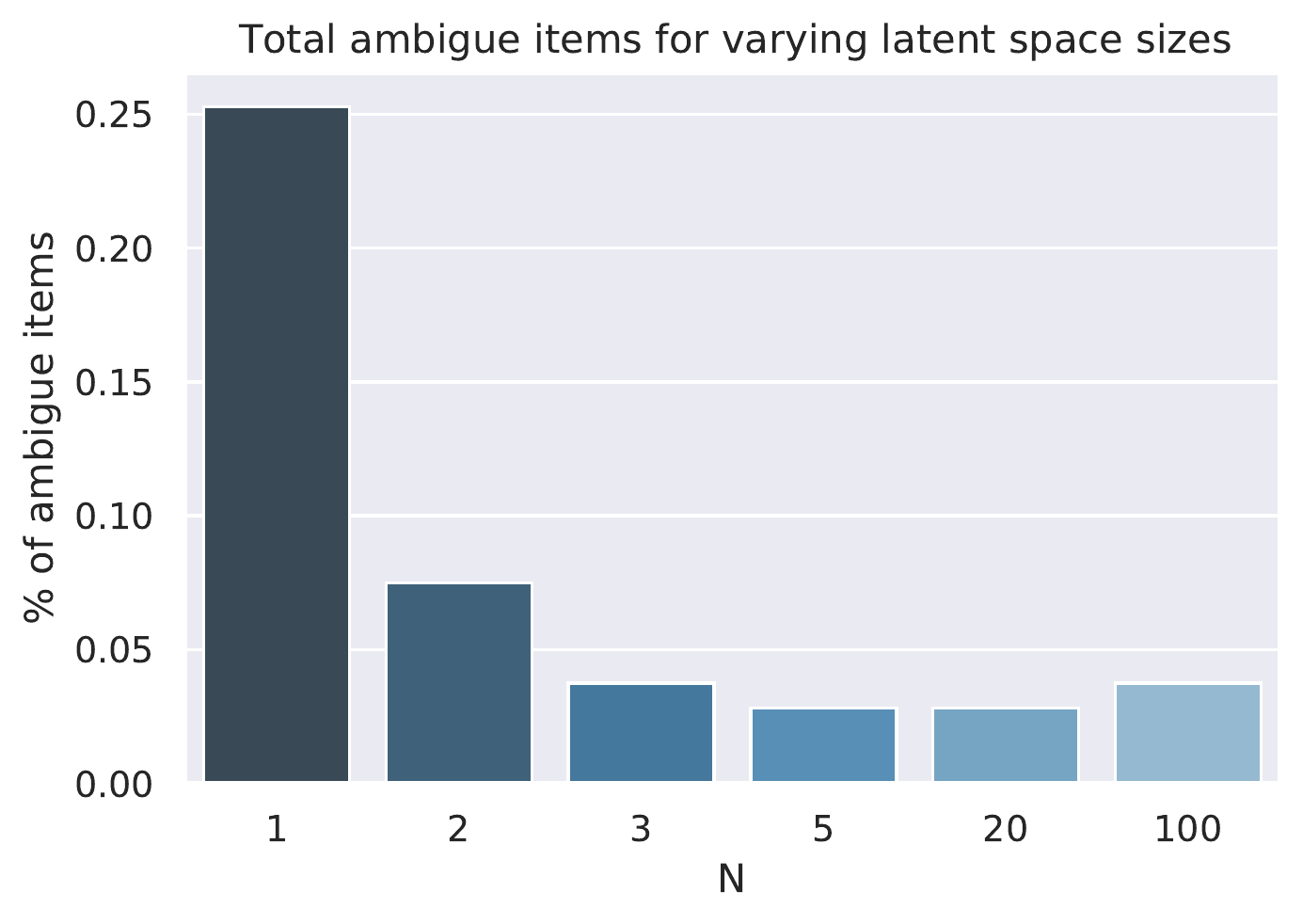}
    \caption{MovieLens dataset}
    \label{fig:ml_amb_perc}
\end{subfigure}
\hspace{1cm}
\begin{subfigure}{.4\textwidth}
  \centering
  \includegraphics[width=\textwidth]{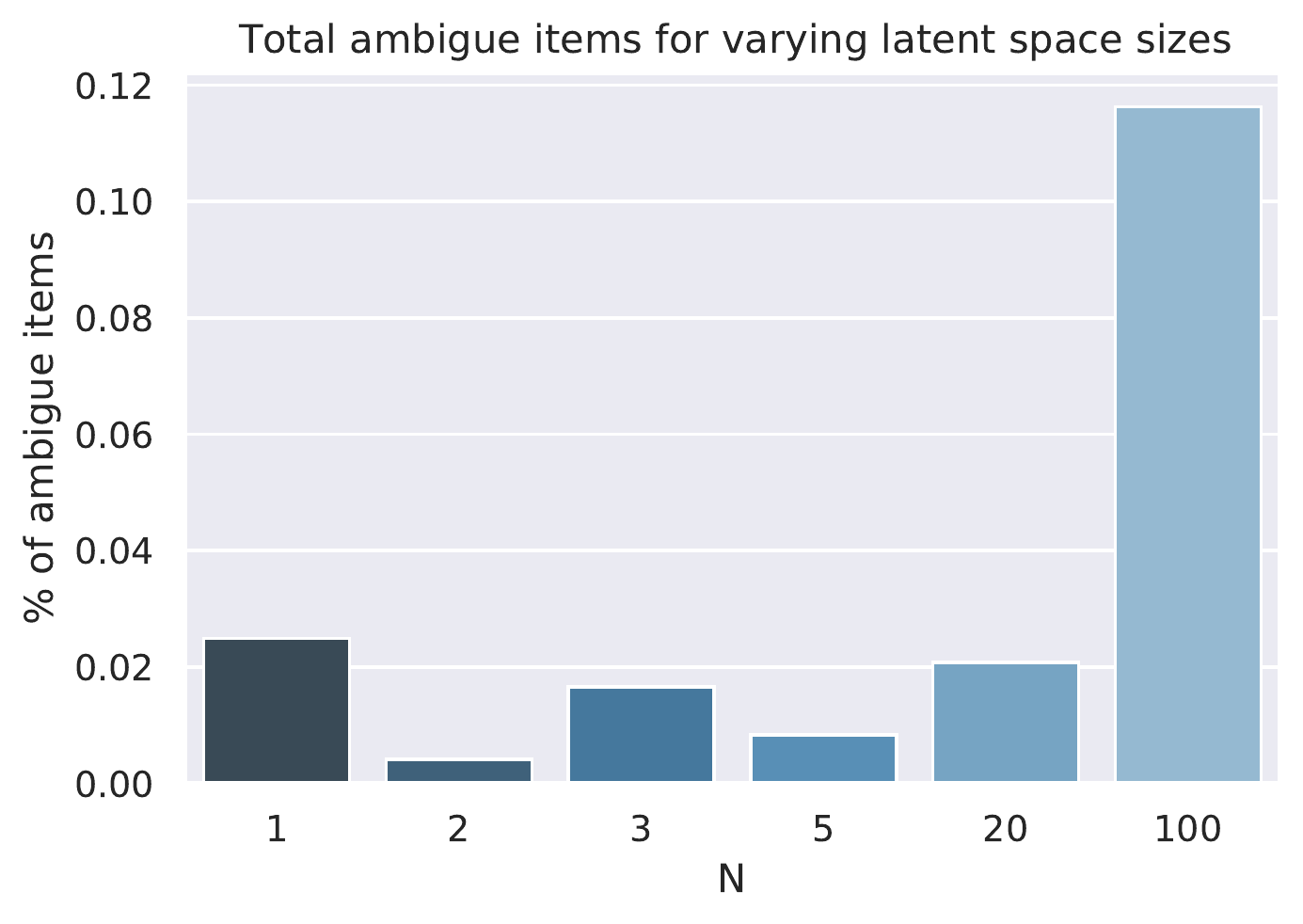}
    \caption{LastFM dataset}
    \label{fig:fm_amb_perc}
\end{subfigure}
\caption{Total percentage of ambigue items ($y$-axis) when comparing the models with different latent space dimension $d$ in the $\epsilon$-level set on their set of available items. We split between different Top-$N$ recommender systems ($x$-axis).}
\label{fig:ambiguity}
\end{figure*}

Shifting our focus, the key concept in the work by \citet{MarxCU19} is the concept of multiplicity. If there are at least two competing models within an error tolerance $\epsilon \geq 0$ the respective problem exhibits multiplicity \citep{MarxCU19}. \citet{MarxCU19} refer to the set of competing models as the $\epsilon$-level set and extend that term to \emph{predictive multiplicity} where two models within the $\epsilon$-level set assign different predictions to a instance $\mathbf{x}_i$ in the training data.

Further, \citet{MarxCU19} propose formal measures for the possibility of multiple competing models (\emph{predictive multiplicity}) in the form of  \emph{discrepancy} and \emph{ambiguity} and define them as:
\begin{definition}[discrepancy]
Maximum number of conflicting predictions between a baseline model and any good model. If the discrepancy is small, near-optimal models (in the $\epsilon$-level set) output similar predictions and vice versa.
\end{definition}
\begin{definition}[ambiguity]
Number of individuals that can be assigned a different prediction by at least one model in the $\epsilon$-level set. 
\end{definition}
While \emph{discrepancy} is an upper bound for the number of predictions that can change, \emph{ambiguity} determines that value for a particular model choice between a set of good models.  
Further, they provide integer programming tools to compute these measures for linear classification problems taking into account all possible models within a certain performance margin \citep{MarxCU19}.

For their experiments, they construct binary classification problems based on the ProPublica COMPAS dataset \citep{machinebias}, the Felony Defendants in Large Urban Counties  dataset \citep{pretrial} and the Recidivism of Prisoners Released in 1994 dataset \citep{recidivism}. The results by \citet{MarxCU19} show that for example on the COMPAS dataset a competing model with only $1\%$ less accuracy can disagree on over $17\%$ of the predictions (discrepancy) and $44\%$ of predictions are vulnerable to model selection (ambiguity). Further, they try to raise awareness that discrepancy and ambiguity should \emph{not} be overlooked when deploying classification systems in highly influential real-world scenarios and should be reported similarly to model statistics such as the test error \citep{MarxCU19}. 
\section{Experimental Setup}

\begin{figure*}
\centering
\begin{subfigure}{.4\textwidth}
  \centering
  \includegraphics[width=\textwidth]{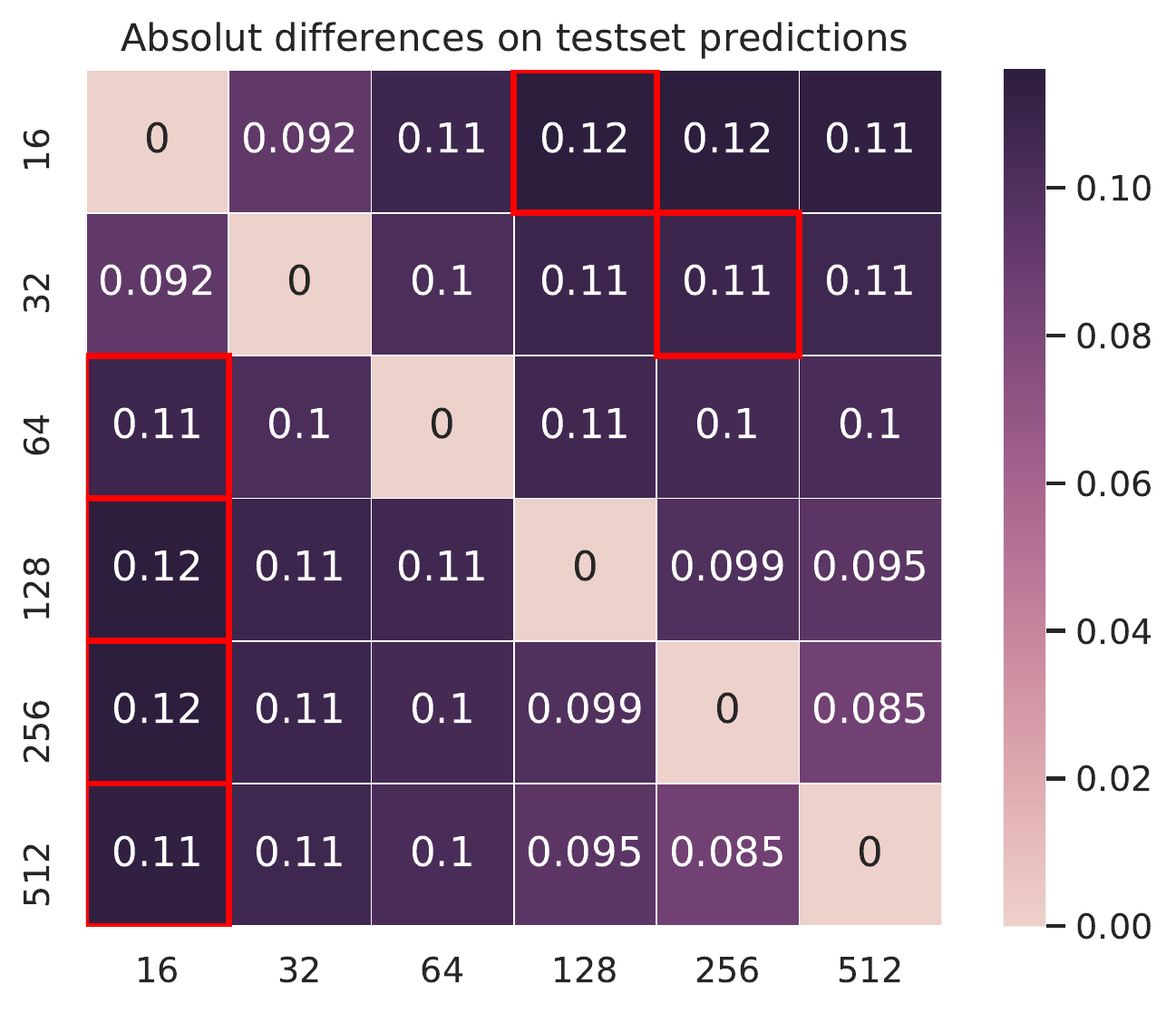}
    \caption{MovieLens dataset}
    \label{fig:ml_pred_diff}
\end{subfigure}
\hspace{1cm}
\begin{subfigure}{.4\textwidth}
  \centering
  \includegraphics[width=\textwidth]{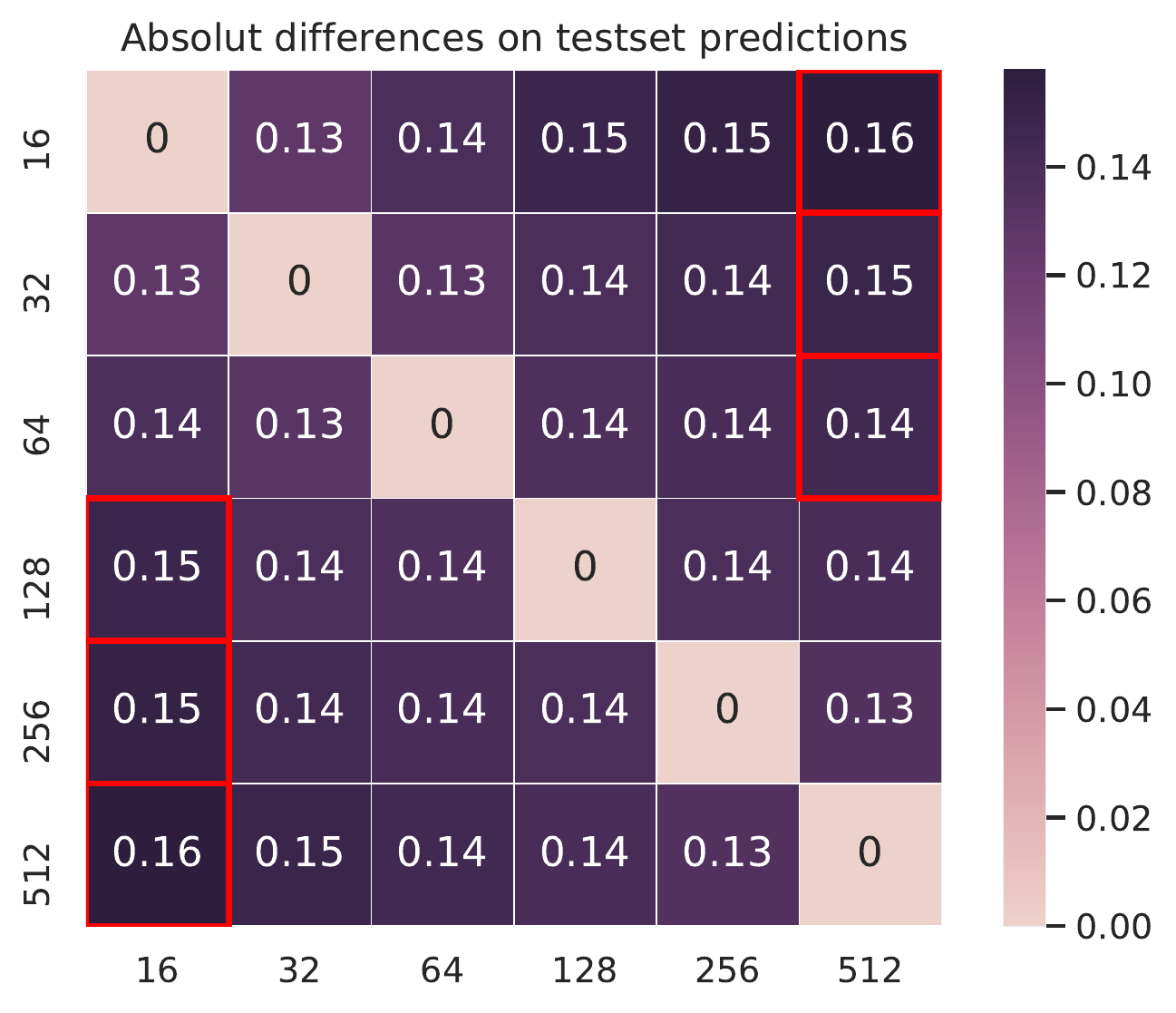}
    \caption{LastFM dataset}
    \label{fig:fm_pred_diff}
\end{subfigure}
\caption{Average absolute rating difference for the testset predictions when comparing any baseline model ($y$-axis) to all models in the $\epsilon$-level set ($x$-axis) on the MovieLens dataset (left) and the LastFM dataset (right). The model with the highest absolute difference to the baseline model (row-wise maximum) is highlighted in red.}
\label{fig:pred_diff}
\end{figure*}

For our experiments, we want to apply the definition of \emph{predictive multiplicity} and evaluate the proposed measures (\emph{discrepancy} and \emph{ambiguity}) from \citet{MarxCU19} on the recommendation system of \citet{DeanRR20}. Therefore, we first need to determine a relevant $\epsilon$-level set. For our experiments, we used the trained models from \citep{DeanRR20}, using the same settings as used in the paper for both, the MovieLens 10M dataset as well as the LastFM 1K dataset. Hence, we use the same values for the number of latent dimensions $d = \{16, 32, 64, 128, 256, 512\}$ while setting the sizes of the recommender set to $N = \{1, 2, 3, 5, 20, 100\}$. As shown in \Cref{fig:pred}, when deploying an $\epsilon = 0.02$ on the Root-Mean-Squared-Error (RMSE), all our trained models with different latent dimensions lie within the $\epsilon$-level set. According to \citet{MarxCU19}, an $\epsilon$-value of $1\%$ represents a conservative default for accuracy based measures and hence a RMSE with $\epsilon = 0.02$ is reasonable when applied in our application scenario. We also verified this by looking at the absolute changes in prediction on the testset in \Cref{fig:pred_diff}. These are also marginally small which verifies our choice of $\epsilon$-level set.

Based on this set of competing models, we can evaluate the \emph{discrepancy} in availability on the testset by choosing a baseline classifier from the $\epsilon$-level set and comparing it to all other models in this set. This way, we can construct $|N|$ comparison matrices for the \textbf{discrepancy of availability with varying $d$} of size $|d| \times |d|$ which for us are $6$ matrices of size $6 \times 6$ for each dataset. Additionally, by also considering $N$ as a model parameter, we can construct comparison matrices of size $|d| \cdot |N| \times |d| \cdot |N|$ for the \textbf{discrepancy of availability with varying $d$ and $N$} which for us is $1$ matrix for each dataset of size $36 \times 36$. By taking the row-wise maximum value, we evaluate the \emph{final discrepancy} for each baseline model choice. With this structure of our experiments, the constructed comparison matrices are symmetric and have diagonal entries with zero values since the set of conflicting elements between the same model is always an empty set.

Note, \citet{DeanRR20} also compare their definitions of \emph{recourse} and \emph{availability} for the different models in our $\epsilon$-level set (cf.~Figures 3-7 and 10-13 in \citep{DeanRR20}), however, they do not include an analysis of the conflicting elements on this set. Hence, we base our analysis of the availability on the conflicting elements when comparing any baseline model and its available set $\mathbb{Y}$ with any model of the $\epsilon$-level set and its available set $\mathbb{X}$ where we define the set of conflicting elements $\mathbb{Z}$ to be the negated intersection as $\mathbb{Z} = \neg \left(\mathbb{Y} \cap \mathbb{X}\right)$. This set includes all items of $\mathbb{Y}$ that are not included in $\mathbb{X}$ and vice versa. We want to stress that we consider the set difference $\mathbb{Z} =  \mathbb{Y} \setminus \mathbb{X}$ as especially relevant metric since these highlight available items in a smaller set which are not available in the broader set which has already progressed further. Therefore, we ensure that the baseline model for $\mathbb{Y}$ has a smaller set size $|\mathbb{Y}|$ than the model of the $\epsilon$-level set for $|\mathbb{X}|$ formalized as $|\mathbb{Y}| < |\mathbb{X}|$. This yields available items which are available in the smaller available set but not in the larger one.

Further, we analyse the \emph{ambiguity} on the respective prediction items by  comparing how often each item falls into a conflicting prediction set $\mathbb{Z}$ for each model of the $\epsilon$-level set. This allows us to compute an average percentage value of \emph{ambigue} items (items that at least fall into one conflicting prediction set) for each element of the recommender set $N$. This yields $|N|$ percentage values of average ambiguity values for the respective recommender system which, in our case, are $6$ values for each dataset.

\section{Results \& Discussion}

We observe some high-level trends where for the same latent space size $d$ for any two Top-$N$ recommender systems with $N_1 < N_2$ the available items of the Top-$N_1$ recommender system are a proper subset of the Top-$N_2$ recommender system. This observation partly motivated our experiments for \Cref{fig:ml_av,fig:fm_av} where we compare the availability for different latent space sizes $d$ for consistent Top-$N$ recommender systems. The results illustrate that, generally speaking, we observe contradicting availability sets when comparing two latent space sizes $d$ that are rather small (e.g.~ $16$, $32$ or $64$). However, for the majority of the Top-$N$ recommender systems (mostly all but $N=1$), contradicting availability values exclusively appear in the upper left quadrant for both the MovieLens and the LastFM dataset. Even if contradicting available items appear, they tend to be a comparably small amount with regard to the overall size of available items. This leaves us with the conclusion that \textbf{when keeping the Top-$N$ recommendation system constant, different models have very little discrepancy in availability}.

Further, when looking at at the ambiguity of the conflicting available items illustrated by \Cref{fig:ambiguity}, we observe a similar pattern. Overall, \textbf{the number of ambigue items for the models in our $\epsilon$-level set with varying latent dimension size $d$ tends to be very low}. For the MovieLens dataset this value ranges from roughly $0.02-0.25\%$ while for the LastFM dataset the range is from $0.005-0.115\%$. Since this value is so low, it is highly unlikely that it will have an overall meaningful impact and therefore can be neglected.

When we expand our $\epsilon$-level set from only considering the latent space dimension $d$ as a model parameter to considering the Top-$N$ recommendation sets in addition to the latent space dimension, we can construct a similar heatmap as seen previously. Now, this new heatmap has shape $36 \times 36$ and is illustrated in \Cref{fig:pred_big}. Here, we observe larger values with a\textbf{ maximum for the MovieLens dataset of $659$ and for the LastFM dataset of $2322$}. The maximum values respectively occur when comparing $N=2,d=32$ (baseline) to $N=20,d=16$ and $N=1,d=256$ (baseline) to $N=100,d=64$. For the MovieLens dataset, $8, 73\%$ of pairings produce a set difference above $1\%$, while $0,79\%$ produce a set difference above $10\%$ with a maximum discrepancy of $16, 35\%$. For the LastFM dataset, $17,78\%$ of pairings produce a set difference above $1\%$, while $4, 29\%$ produce a set difference above $10\%$ with a maximum discrepancy of $29, 19\%$.
There seems to be no direct correlation between the discrepancy and the difference in set size.

\section{Outlook}

As one can see, there are a lot of possible aspects one can analyse when looking at a recommendation system. Obviously, it is impossible to cover all possible choices of such degrees of freedom and hence our analysis is mostly biased towards the availability (and therefore inherently the recourse) in recommendation systems, since we identify this as a key challenge and very critical point across different models. Further, additional analysis can be done in the future on comparing the \emph{discrepancy} and \emph{ambiguity} for \textbf{recommendations for the user cold start}, \textbf{cost of recourse} or \textbf{recommendations based on a fixed history} on models with different latent space size $d$.

\printbibliography

\end{document}